\documentclass[letterpaper, 10 pt, conference]{ieeeconf}  

\IEEEoverridecommandlockouts                              

\usepackage{multicol}
\usepackage[bookmarks=true]{hyperref}
\usepackage{bm}
\usepackage{amssymb}
\usepackage{threeparttable}
\usepackage{booktabs}
\usepackage{wrapfig}
\usepackage{amsmath}
\usepackage{wrapfig}
\usepackage{lipsum}
\usepackage{siunitx}
\usepackage{graphicx}
\usepackage{url}
\usepackage{cleveref}
\usepackage{cite}
\usepackage{times}
\usepackage[table,xcdraw]{xcolor}
\usepackage{multirow}
\usepackage{color}
\usepackage{siunitx} 
\usepackage{xcolor}
\usepackage{soul}
\title{\LARGE \bf
SUBTA: A Framework for Supported User-Guided Bimanual Teleoperation in Structured Assembly
}

\author{Xiao Liu$^{1}$, Prakash Baskaran$^{1}$, Songpo Li$^{1}$, Simon Manschitz$^{2}$, Wei Ma$^{2}$, Dirk Ruiken$^{2}$, and Soshi Iba$^{1}$
\thanks{$^{1}$Authors are with Honda Research Institute USA, San Jose, CA 95134 USA.}
\thanks{$^{2}$Authors are with Honda Research Institute Europe GmbH, 63073 Offenbach am Main, Germany.}
\thanks{Correspondence: {\tt\small songpo\_li@honda-ri.com}}
}

\begin{document}

\maketitle
\thispagestyle{empty}
\pagestyle{empty}


\begin{abstract}

    In human-robot collaboration, shared autonomy enhances human performance through precise, intuitive support. Effective robotic assistance requires accurately inferring human intentions and understanding task structures to determine optimal support timing and methods. In this paper, we present \textsc{SUBTA}, a supported teleoperation system for bimanual assembly that couples learned intention estimation, scene-graph task planning, and context-dependent motion assists. We validate our approach through a user study (\(N{=}12\)) comparing standard teleoperation, motion-support only, and \textsc{SUBTA}. Linear mixed-effects analysis revealed that \textsc{SUBTA} significantly outperformed standard teleoperation in position accuracy ($p<0.001$, $d=1.18$) and orientation accuracy ($p<0.001$, $d=1.75$), while reducing mental demand ($p=0.002$, $d=1.34$). Post-experiment ratings indicate clearer, more trustworthy visual feedback and predictable interventions in \textsc{SUBTA}. The results demonstrate that \textsc{SUBTA} greatly improves both effectiveness and user experience in teleoperation.
\end{abstract}

\section{INTRODUCTION}

Teleoperation refers to operating a machine or robot from a distance, often to allow humans to perform tasks remotely or in hazardous environments. It has proven invaluable in domains such as bomb disposal, underwater exploration, and space missions, where keeping human operators out of danger is paramount~\cite{cui2003review}. In the manufacturing context, robotic teleoperation enables human workers to execute complex and precise tasks (for example, intricate assembly, welding, or maintenance operations) from a safe location~\cite{zheng2024design}. This approach not only minimizes risks to workers in dangerous or confined factory settings, but also leverages human dexterity and decision-making for tasks that are too variable or difficult to fully automate. Manufacturing teleoperation can thus combine the best of human skill and robotic precision, improving safety and flexibility on the factory floor~\cite{zheng2024design,gonzalez2021advanced}. However, directly manipulating a robotic arm through a standard interface can be challenging due to differences in the robot’s kinematics and sensing compared to a human’s, often necessitating extensive training and robotics expertise~\cite{manschitz2022shared}. Moreover, pure teleoperation demands continuous operator attention and a fast, stable communication link, since the human must control every motion in real-time~\cite{senft2021task}. These factors make generic teleoperation interfaces cumbersome for manufacturing tasks, especially for operators who are not robotics specialists~\cite{senft2021task}.

\begin{figure}[t]
  \centering
  \includegraphics[width=0.8\columnwidth]{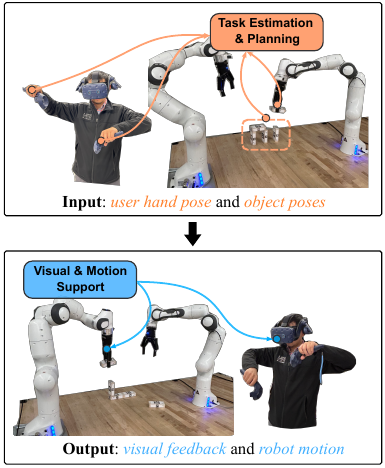}
  \caption{
  \textbf{SUBTA}: The proposed Supported Teleoperation framework assists users in assembly tasks by (i) estimating the task and generating a plan, and (ii) providing real-time support through visual feedback and motion-level corrections.
}
  \label{fig:teaser}
  \vspace{-0.15in}
\end{figure}

To address these challenges, researchers and engineers are turning to task-specific supported teleoperation systems. The central idea is that by supporting the human with intelligent interfaces and autonomy, even non-expert users can perform complex tasks more easily and efficiently than with a one-size-fits-all teleoperation setup~\cite{zheng2024design,akita2024improved}. For example, in \Cref{fig:teaser}, our system assists a user during a block-assembly task, which is representative of many industrial assembly operations~\cite{kulshrestha2023structural,kiyokawa2023difficulty}. It estimates user intent, models task state with a scene graph~\cite{jiang2024roboexp}, and provides motion support accordingly. Such task-specific support dramatically lowers the barrier to using robots in manufacturing, enabling users to leverage robots for intricate tasks without needing to micromanage every action. 

In this paper, we propose a \textbf{S}upported \textbf{U}ser-Guided \textbf{B}imanual \textbf{T}eleoperation system for \textbf{A}ssembly tasks (\textsc{SUBTA}), extend task-specific support in teleoperation by incorporating task state estimation and graph-based task planning. This enables the robot to understand the task structure in 3D space and assist the user through both visual feedback and motion-level corrections. Our main contributions are summarized as follows:
\begin{itemize}
    \item Shared autonomy system that integrates three levels of assistance: task understanding and intention estimation, task planning, and low-level motion support.
    
    \item Scene graph representation that encodes spatial relationships in structured assembly tasks for task state estimation and planning.

    \item Executing a set of gated motion behaviors for grasp and placement, coordinated by a behavior controller to deliver support ``when and how'' it is needed
    
    \item Comprehensive experimental validation with a user study demonstrating the effectiveness of task-specific support in improving teleoperation performance. 
    
    \item SUBTA reduced cognitive load (6.2 $\rightarrow$ 3.4) while maintaining a 75\% success rate, enabling non-experts to perform structured, multi-step assemblies more effectively with nearly twofold higher accuracy.
\end{itemize}

\begin{figure*}[t]
  \centering
  \includegraphics[width=1.9\columnwidth]{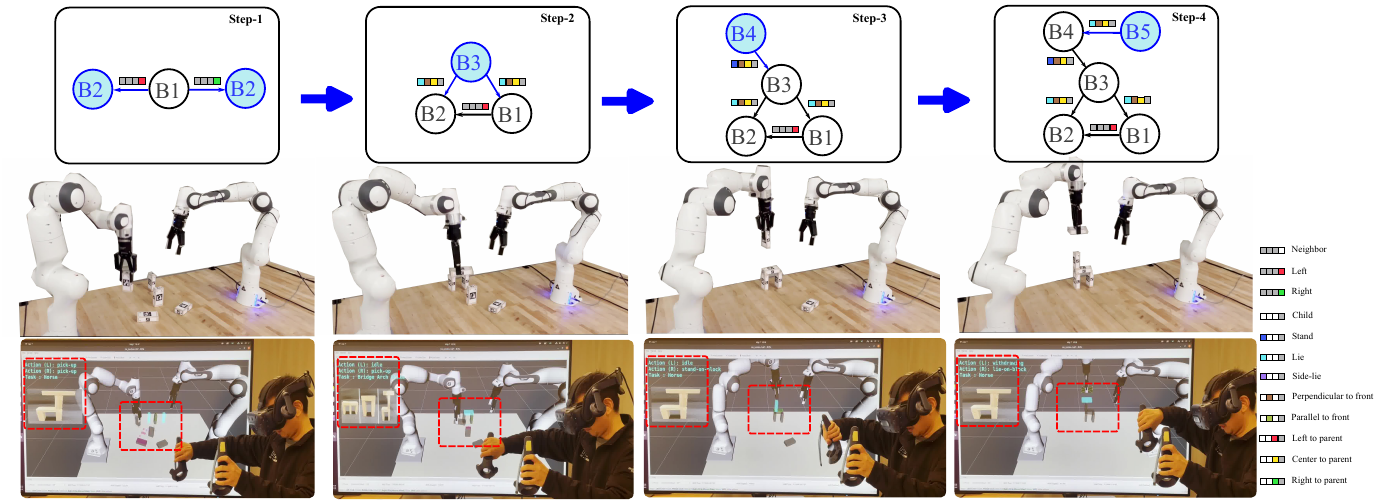}
  \caption{\textbf{Supported teleoperation system}: The user teleoperates the robot to assemble the ``horse'' structure. 
The \textbf{top row} shows the scene graph encoding spatial relationships for task monitoring and planning. 
The \textbf{middle row} depicts the robot successfully executing the block assembly. 
The \textbf{bottom row} presents the digital twin environment, where task estimation and planning ((highlighted in red) are visualized to guide the user.
}
  \label{fig:architecture}
  \vspace{-0.4cm}
\end{figure*}

\section{Related Work}
Purely manual teleoperation is widely acknowledged to have significant limitations~\cite{rea2022still}. Operators must continuously control every robot motion, resulting in high cognitive and physical demands, increasing susceptibility to fatigue, and potentially introducing human errors, particularly during prolonged operations~\cite{lopez2022assessing}. Communication delays or network instabilities exacerbate these issues, as lags in visual feedback or control signals complicate tasks that require precise timing or delicate coordination~\cite{farajiparvar2020brief}. Additionally, disparities between the human operator's embodiment and the robotic system—such as differences in scale, degrees of freedom, or sensory inputs—often hinder intuitive control, particularly for complex robotic arms~\cite{dragan2013teleoperation}.

To address these fundamental limitations, several streams of research have emerged, focusing on improving usability, reducing cognitive load, and enhancing performance in teleoperation systems.

\emph{High-level task interfaces} represent an early and significant advancement in teleoperation. Rather than relying on low-level, joint-by-joint control, these interfaces allow users to issue commands at a task or semantic level~\cite{meng2023virtual}. Researchers have proposed intuitive authoring tools tailored for light manufacturing scenarios, showing that novice users can accomplish complex assembly jobs more efficiently compared to traditional joystick-based methods~\cite{wang2023built, aoyama2024asynchronously}. These studies consistently indicate that abstracting teleoperation tasks into high-level semantic commands such as ``pick-and-place" significantly enhances usability, particularly for users without extensive robotic training.

\emph{Shared autonomy} has gained traction by introducing real-time robotic assistance guided by intent inference. For instance, Manschitz et al.~\cite{manschitz2022shared,manschitz2025icra} provided visual and haptic feedback and automated trajectory corrections to help operators accurately perform pick-and-place tasks, especially compensating for common depth-perception errors while maintaining the operator's central role. Similarly, Owan et al.~\cite{owan2022dynamic} developed dynamic autonomy systems capable of adaptively shifting control between automated execution and human guidance, particularly beneficial in confined-space tasks. These approaches emphasized precision and efficiency while preserving human agency and decision-making capabilities during teleoperation.

Further advancements include \emph{augmented reality} (AR) and \emph{digital twin} technologies, designed to further reduce cognitive load and enhance the operator’s situational awareness by visually contextualizing tasks and pre-validating robotic actions. Lu et al.~\cite{lu2020digital} successfully integrated AR headsets with digital twins representing virtual replicas of the robotic work environment. Operators issue high-level commands within this virtual context, which immediately evaluates command feasibility and prevents inappropriate or impossible actions. By providing operators with an informed and interactive task environment, these systems transform robots into collaborative partners that inherently understand operational constraints and offer immediate, actionable feedback~\cite{zhang2024enabling}.

\section{Task-specific Teleoperation Framework}

The goal of our framework is to support human operators in completing manipulation tasks. In this work, we focus on pick-and-place actions within structured assembly tasks, which require precise object handling to form spatial configurations.
The system consists of four main components:
1) Task and Intention Estimation Module: This module takes as input the human operator’s motion and the current poses of objects. It estimates the user’s intention, including the task being performed and the specific action being taken.
2) Task Planning Module: The current task is modeled as a scene graph to represent the task state accurately. This allows the system to track progress and determine the remaining steps needed to complete the task.
3) Behavior Controller: A state machine that, based on the current context, manages the activation and deactivation of low-level assistance during grasping and placement. 
It automatically attaches the object to the operator’s hand when grasping and snaps it to the target location when the hand is near the desired placement.


\subsection{Task and Intention Estimation}
Given $\text{SE(3)}$ poses with position and quaternion for orientation of both hands 
$\mathbf{H}_{t}\!\in\!\mathbb{R}^{6\times2}$ and $n$ blocks  
$\mathbf{B}_{t}\!\in\!\mathbb{R}^{6\times n}$,  
the module predicts the current task label $\mathbf{l}_{t}$, and left and right‐hand actions $\mathbf{a}_{t}$. The estimation model encodes the human hands $\mathbf{H}_{t}$, with \textit{tAPE} positional encoding~\cite{ConvTran2023}, outputs seven 32-dim embeddings (i.e., one embedding per entity) that serve as node features for a graph neural network (GNN). We adapt the structure from the HAR-Transformer~\cite{dirgova2022wearable}. The operator's intention is inferred from evolving spatial relations among entities, modeled as a graph with nodes $\mathbf{V} = \{v_l, v_r, v_{b_1}, \ldots, v_{b_n}\}$, representing the hands and blocks. The adjacency matrix $\mathbf{A}_t$ is dynamic, adapting over time as interactions progress. $\mathbf{A}_t$ is computed using a self-attention mechanism \cite{vaswani2017attention} applied to the input node features. Attention scores are scaled, normalized via softmax, and symmetrized by averaging with their transpose:
\begin{equation}
 \begin{array}{l}
    \textit{Attn}(Q, K, V) = \textit{softmax} \left( \frac{QK^T}{\sqrt{d_k}} \right) , \quad
    \textit{softmax}(x_i) = \frac{e^{x_i}}{\sum e^{x_i}} \\
    \mathbf{A}_t = (\textit{Attn} \, + \, \textit{Attn}^T)/2 \:,
\end{array}
\end{equation}
\noindent where, $Q, K, V$ are the input features and $d_k$ their dimensionality. The resulting $\mathbf{A}_t$ captures learned, symmetric spatial relationships and is optimized end-to-end.
The GNN extracts spatial features by applying graph convolutions across $K$ layers, each followed by an activation (except the final layer). The layer-wise computation is:
\begin{equation}
 \begin{array}{l}
     \boldsymbol{Z}_t^{(k)} = \mathit{f_k} \, (\mathbf{A}_t, \boldsymbol{F}_t^{(k-1)}) \:,\\
     \boldsymbol{Z}_t^{(k)} = \mathbf{A}_t * \boldsymbol{F}_t^{(k-1)} * \boldsymbol{W}^{(k-1)} \\
     \boldsymbol{F}_t^{(k)} = \alpha_k \, (\boldsymbol{Z}_t^{(k)}) \:,
\end{array}
\end{equation}
\noindent where $\mathbf{A}_t \in \mathbb{R}^{(n+2) \times (n+2)}$ is the adjacency matrix, and $\boldsymbol{F}^{(0)}$ is the GNN's input node feature layer. Each layer outputs $\boldsymbol{F}_t^{(k)} \in \mathbb{R}^{(n+2) \times d_{k}}$, with learnable weights $\boldsymbol{W}^{(k-1)} \in \mathbb{R}^{d_{k-1} \times d_{k}}$. 
The final GNN output $\mathbf{F}^{(2)}$ is flattened, passed through a 32-unit ReLU layer, and forms a feature vector $\mathbf{g}$.
This vector feeds three heads that predict (i) the task~(sigmoid), (ii) left-hand action, and (iii) right-hand action. The action heads concatenate $\mathbf{g}$ with velocity embeddings from a three-layer HAR-Transformer~\cite{dirgova2022wearable}. The training is conducted with Adam (lr = $10^{-3}$), using class-weighted binary cross-entropy for the task head and focal loss for the action heads to address class imbalance. Inputs are 3 sec windows (20 Hz) with a 1 sec stride.

\subsection{Task Planning}
\begin{figure}[t]
  \centering
  \includegraphics[width=0.9\columnwidth]{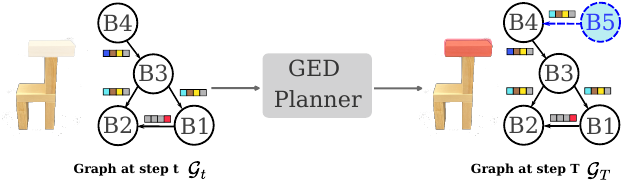}
  \caption{
  \textbf{Task Planner}: the planner computes the most likely target pose for the next block based on the graph edit distance~(GED) and adds the corresponding node to the graph.}
  \label{fig:GED}
  \vspace{-0.10in}
\end{figure}

Scene graphs have been shown to effectively capture diverse object relationships in manipulation tasks~\cite{zhai2024sg}. 
In this work, we extend this idea by employing scene graphs to represent the geometric structure of assembly tasks. Given
$\mathbf{B}_{t}\!\in\!\mathbb{R}^{7\times n}$ as the set of $\text{SE(3)}$ poses used in one assembly. We design a heuristic, rule-based algorithm $h$ to extract the pairwise spatial relationships between blocks. When constructing the scene graph, we create edges only between blocks that are in contact or at the same height. We represent the scene graph as:

 \begin{equation}
 \begin{array}{l}
     \mathcal{G}_t = (\pmb{V}_t, \pmb{E}_t, \pmb{\epsilon}_t, h),\\
     h: \pmb{\epsilon}_t \rightarrow \{\{v_n , v_m \} | v_n , v_m \in \pmb{V}_t\: \text{and}\: h(v_n , v_m)\neq \text{None}\}.\\
\end{array}
\end{equation}
Each node $v_t \in \pmb{V}_t$ carries the raw pose of its block.
When the operator begins manipulating a new block, a new node for that block is appended to the graph, and all incident edges~$e_t \in \pmb{E}_t$ together with its attributes $\epsilon_t \in \pmb{\epsilon}_t$ are generated by the heuristic $h$ before being added. 
Edge attributes, defined in Table~\ref{tab:attr_labels}, encode the spatial relationships between blocks during assembly. 
For example, in \Cref{fig:GED}, block $B_5$ and block $B_4$ are connected with the attribute $\epsilon_t = [2,2,1,0]$, 
indicating that $B_5$ is placed in lying pose and centrally on top of $B_4$ and oriented perpendicular to its front face.

\begin{table}[t]
  \scriptsize
  \setlength{\tabcolsep}{1.2pt}
  \caption{Discrete spatial‐relation edge attributes.}
  \label{tab:attr_labels}
  \begin{tabular}{c c c c c}
    \toprule
    \textbf{Label} & \textbf{Ori.\ w.r.t.\ Parent} & \textbf{Ori.\ w.r.t.\ Front} & \textbf{Pos.\ w.r.t.\ Parent} & \textbf{Pos.\ w.r.t.\ Neighbor} \\ 
    \midrule
    0 & --      & --            & --     & --    \\
    1 & Stand   & Parallel       & Center & Left  \\
    2 & Lie     & Perpendicular  & Left   & Right \\
    3 & Side‐lie& --             & Right  & --    \\
    \bottomrule
   \multicolumn{5}{l}{Ori.\ = Orientation, Pos.\ = Position}
  \end{tabular}
  \vspace{-0.25in}
\end{table}

\begin{table*}[t]
  \scriptsize
  \setlength{\tabcolsep}{8.5pt}  
  \caption{Context-dependent behaviours for pick–place motion support.}
  \label{tab:behaviours}
  \begin{tabular}{llllll}
    \toprule
    \# & \textbf{Behaviour} & \textbf{Trigger} & \textbf{User control} & \textbf{System autonomy / constraints} & \textbf{Feedback} \\
    \midrule
    1 & Approach Object & Intention to pick \& object detected & 6-DoF free motion & Limit joint/speed for safety; & - \\
    2 & Snap to Object  & Hand to object dist.\,$<\!\delta_1$ & Frozen & Auto-drive EE to grasp pose & Object highlight\\
    3 & Align w/ Object  & Snap to object done & Nullspace motion & Collision avoidance in grasp manifold & -\\
    4 & Grasp Object     & Button press & EE locked & Close gripper & Haptic click \\
    5 & Align w/ Surface & Object grasped, on plane & on-plane transformation & Keep contact; suppress lift/tilt & Plane highlight \\
    6 & Unsnap Surface   & Controller lifted $>\!\delta_2$ & - & Safe lift motion & - \\
    7 & Approach Surface & Unsnap surface done & 6-DoF free motion & Limit joint/speed for safety & - \\
    8 & Snap to Surface  & Hand to plane dist.\,$<\!\delta_3$ & Frozen & Auto-drive EE to plane, Keep contact & Plane highlight\\
    9 & Release Object   & Finger‐opend & EE locked & Open fingers; release & - \\
    \bottomrule
    \multicolumn{6}{l}{End-effector (EE), distance (dist.), $\delta_1, \delta_2$ and $\delta_3$ are adjustable thresholds.} \\
  \end{tabular}
  \vspace{-0.15in}
\end{table*}

The task planning is achieved by using graph edit distance~(GED). The planning and next step target prediction can be combined to automatically complete the task at any stage of a trial. We define a set of graph edit operations $\pi = \{o_1, o_2, o_3\}$, where $o_1$ refers to add-delete-modify any node $v_i$, $o_2$ refers to add-delete-modify any edge $e_i$, and $o_3$ refers to add-delete-modify any edge attributes $\epsilon_i$. Given a current scene graph $\mathcal{G}_{t}$, and the end scene graph $\mathcal{G}_{T}$, we use the algorithm
 \begin{equation}
 \begin{array}{l}
d_{\text{GED}}(\mathcal{G}_{t},\mathcal{G}_{T})=\min_{\pi \in \mathcal{P}(\mathcal{G}_{t},\mathcal{G}_{T})}\sum_{\,o \in \pi} c(o),
\end{array}
\end{equation}
where $\mathcal{P}(\mathcal{G}_{t},\mathcal{G}_{T})$ is the set of edit paths that transform $\mathcal{G}_{t}$ into $\mathcal{G}_{T}$, $c(o)\!\ge\!0$ is the cost assigned to operation~$o$. \Cref{fig:GED} illustrates the procedure for the ``horse'' assembly task. At each time step the planner constructs the current scene graph 
$\mathcal{G}_{t}$, compares it with the goal graph $\mathcal{G}_{T}$, and produces both the graph-edit distance
$d_{\mathrm{GED}}$ and the next intermediate graph on the optimal edit path. The goal graph is selected by the task estimation module, where one assembly can have multiple goal graphs.
In the right panel of \Cref{fig:GED}, block $B_{5}$ (highlighted in blue) is selected as the next target.
After a target node $v_{i}$ is selected, the algorithm retrieves the
corresponding parent and adjacent nodes using simply Breadth-first search. A rule-based heuristic then
estimates the required pose for $v_{i}$ based on the edge attributes, and renders this pose in the
digital-twin environment, providing real-time guidance to the operator.

\subsection{Motion Support Behaviors}
We decompose the overall pick-and-place routine into nine context-dependent behaviors. 
Transitions between behaviors are governed by inferred operator intention, hand--object distance, and motion feasibility. 
This modular design enables graduated support: full manual control is retained during coarse motions, while autonomous corrections are introduced in precision phases. 
Details of each behavior are summarized in \Cref{tab:behaviours}. 
Visual feedback to the operator includes highlighted objects during snapping, emphasized block surfaces for alignment, and snapping indicators to surfaces during placement.


\section{Experiments Design}

\subsection{Experimental setup}
\label{sec:exp_set_up}
Manipulation sequences were collected from humans performing block assembly tasks 
through teleoperation of a bimanual robotic system in a virtual reality environment~\cite{belardinelli2022intention, manschitz2022shared}. 
Following the setup described in~\cite{baskaran2025explainable}, participants executed eight distinct block assembly tasks 
within a virtual scene rendered in Rviz and displayed using an HTC Vive Pro Eye headset. 
The scene consisted of a table with five identical wooden blocks available for assembly. 
Block poses were reliably tracked using ArUco markers detected by three Intel RealSense D435 cameras.
\begin{figure}[tpb]
  \centering
  \includegraphics[width=0.8\columnwidth]{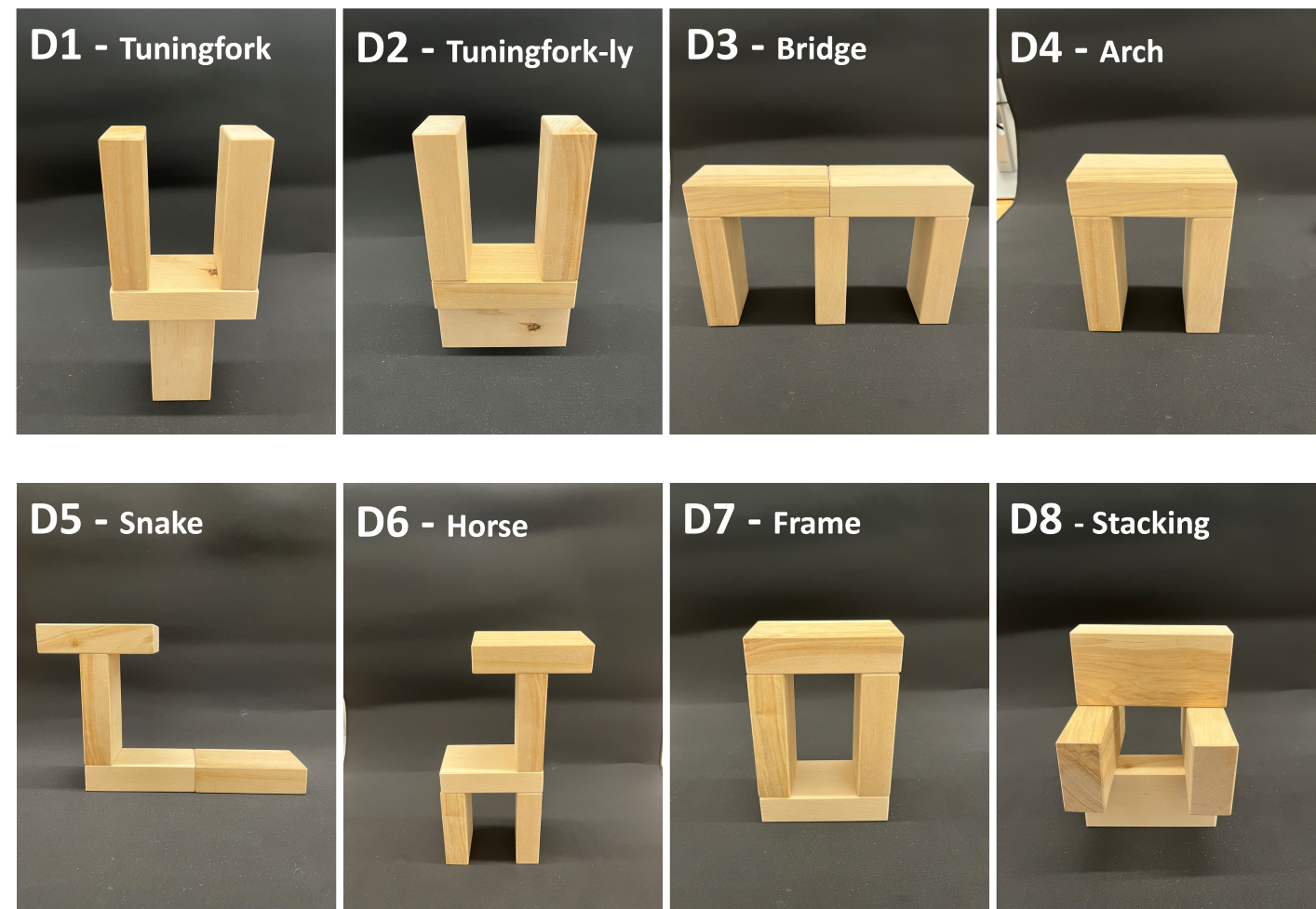}
  \caption{Eight Block assembly tasks.}
  \label{fig:block_assembly}
  \vspace{-0.1in}
\end{figure}
\begin{table}[tpb]
\centering
\caption{Comparison of F1-scores (\% mean $\pm$ std.\@ dev.)}
\scalebox{0.9}{
\label{tab:f1_comparison}
\begin{tabular}{lccc}
\toprule
\textbf{Model} & \textbf{Action (L)} & \textbf{Action (R)} & \textbf{Task} \\
\midrule
Naive CNN        & 73.53 (6.40) & 72.34 (5.61) & 77.58 (4.38) \\
Hierarchical InEs & 43.97 (10.97) & 38.93 (8.90) & 49.43 (8.62) \\
HAR-Transformer  & 81.49 (4.55) & 79.95 (4.39) & \textbf{90.12 (3.71)} \\
\rowcolor[HTML]{E6F4FF} 
Ours             & \textbf{81.73 (4.47)} & \textbf{80.48 (4.84)} & 89.82 (3.75) \\
\bottomrule
\end{tabular}}
\vspace{-0.15in}
\end{table}

Dataset for task estimation module was collected from 495 teleoperation demonstrations form various block assembly tasks. Nine distinct actions defined by end-effector movements were performed to complete these tasks: Idle, Pick-up, Withdraw, Stand, Lie, Side-lie, Stand-on-block (Stand-OB), Lie-on-block (Lie-OB), and Side-lie-on-block (Side-lie-OB). Each teleoperated demonstration required executing an entire task through action sequences selected freely by the participants. The initial positions and orientations of the blocks were randomized prior to each demonstration. Left and right actions were labeled independently according to their start and end times, while task labels were multi-labeled due to partial or complete overlaps, such as the Arch task overlapping with Bridge and Horse tasks (see \Cref{fig:block_assembly}). The dataset consists of totaling 480.4 minutes, and includes recordings of~SE(3) poses of five wooden blocks, end-effector poses, and egocentric videos of teleoperators. 

\begin{figure*}[t]
  \centering
  \includegraphics[width=1.9\columnwidth]{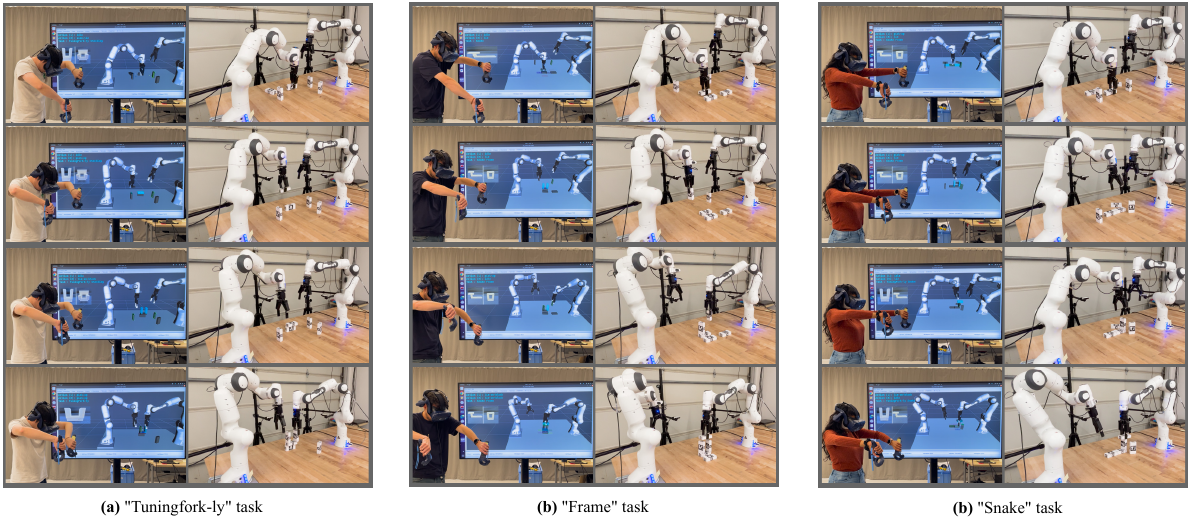}
  \caption{Examples of participants performing three different assembly tasks using the SUBTA system during the user study. 
The supported workspace is projected on the left with the user included, while the actual robot workspace is shown on the right. }
  \label{fig:subject_demo}
  \vspace{-0.4cm}
\end{figure*}

\subsection{Task Estimation Evaluation}

\subsubsection{Baselines} The proposed GNN networks are benchmarked against three baselines: i) Naive CNN architecture, a three-layered 1-D convolutional network, ii) \textit{Hierarchical InEs}~\cite{cai2024hierarchical}, and iii) State-of-the-art \textit{HAR-Transformer}~\cite{dirgova2022wearable}. The latter two models were adapted appropriately. 

\subsubsection{Evaluation} The intention algorithms were validated using the \textit{leave-one-subject-out} cross-validation scheme, where the mean F1-score is reported by training the algorithms repeatedly on all, but one participant's data and validated using the left-out participant's data~\cite{hastie2009elements}.

Table~\ref{tab:f1_comparison} shows that the proposed model substantially outperforms the naive CNN and hierarchical baselines, achieving performance on par with or exceeding the state-of-the-art HAR-Transformer for task and action prediction. Across models, left-hand predictions are 2–4\% more accurate than right-hand predictions, reflecting the predominance of right-handed participants whose left-hand activity is skewed toward the easier-to-predict Idle class. Finally, task predictions consistently surpass action predictions, indicating that short-term action prediction is inherently more difficult due to its transient and dynamic nature. We then integrate this module into the SUBTA system for more comprehensive user studies.

\subsection{Subject Study}
\subsubsection{The goal of the study} 
The goal of this study is to evaluate the accuracy and effectiveness of the proposed 
\textsc{SUBTA} system against standard teleoperation in robotic assembly tasks. 
We compare three modes: 
\begin{enumerate}
    \item \textbf{Standard teleoperation} (M1): direct retargeting with no additional support.
    \item \textbf{Motion-support only} (M2): an ablation of \textsc{SUBTA} providing only general motion assistance. 
    \item \textbf{SUBTA} (M3): with both task-specific and motion support.
\end{enumerate}
System performance is evaluated based on task success rate, completion time, and target pose accuracy. In addition to these objective metrics, participants provide subjective feedback through the NASA Task Load Index (NASA-TLX) and the System Usability Scale (SUS)~\cite{lewis2018system}. After completing all modes, participants also complete another survey rating their overall experience with each interface.

\begin{table}[t]
\centering
\caption{
Linear mixed-effects model results across the three modes}
\scalebox{0.93}{
\label{tab:hypothesis_pvalues}
\begin{tabular}{lp{1.3cm} p{1.3cm} p{1.3cm} p{1cm}}
\toprule
\textbf{Hypothesis}          & \textbf{M3 vs. M2}   & \textbf{M2 vs. M1}   & \cellcolor[HTML]{E6F4FF}\textbf{M3 vs. M1}   & \textbf{Supported} \\
\midrule
$\pmb{H_1 a}$: Pos. err.      & * & ** & \cellcolor[HTML]{E6F4FF}** & Yes \\
$\pmb{H_1 b}$: Ori. err.      & ** & ** & \cellcolor[HTML]{E6F4FF}** & Yes \\
$\pmb{H_3}$: Mental      & NS & ** & \cellcolor[HTML]{E6F4FF}** & Yes \\
$\pmb{H_4}$: SUS            & NS & * & \cellcolor[HTML]{E6F4FF}* & Yes \\
\bottomrule
\multicolumn{5}{l}{NS (not significant); ** $p <0.01$; * $0.01< p <0.05$.} \\
\end{tabular}}
\vspace{-0.15in}
\end{table}

\begin{table*}[t!]
\centering
\caption{Per-task and overall summaries (mean $\pm$ std) by mode. Best value per task is bolded.}
\label{tab:mode_summary_per_task_multirow}
\begin{tabular}{l c c c c c c}
\toprule
\textbf{Task} & \textbf{Mode} & \textbf{Time} (sec) & \textbf{Success Rate} & \textbf{Progress to Complete} & \textbf{Orientation Error} (deg) & \textbf{Position Error} (meter) \\
\midrule
\multirow{3}{*}{``Tuningfork-ly''} & M1 & 74.56 $\pm$ --           & 11.1\% & 38.9\% $\pm$ 25.3\% & 3.710 $\pm$ --      & 0.022 $\pm$ -- \\
                                        & M2 & \textbf{71.02} $\pm$ --   & 12.5\% & 43.8\% $\pm$ 29.1\% & 2.540 $\pm$ --      & 0.013 $\pm$ -- \\
                                        & \cellcolor[HTML]{E6F4FF}M3 & \cellcolor[HTML]{E6F4FF}87.05 $\pm$ 38.26         & \cellcolor[HTML]{E6F4FF}\textbf{22.2\%} & \cellcolor[HTML]{E6F4FF}\textbf{50.0\% $\pm$ 33.1\%} & \cellcolor[HTML]{E6F4FF}\textbf{0.920 $\pm$ 0.382} & \cellcolor[HTML]{E6F4FF}\textbf{0.011 $\pm$ 0.006} \\
\addlinespace
\multirow{3}{*}{``Arch''}          & M1 & 64.14 $\pm$ 13.51         & 62.5\% & 74.9\% $\pm$ 34.7\% & 5.288 $\pm$ 3.310   & 0.028 $\pm$ 0.018 \\
                                        & M2 & 71.32 $\pm$ 15.13         & 85.7\% & 95.1\% $\pm$ 12.9\% & 2.568 $\pm$ 2.032   & 0.013 $\pm$ 0.004 \\
                                        & \cellcolor[HTML]{E6F4FF}M3 & \cellcolor[HTML]{E6F4FF}\textbf{57.86 $\pm$ 7.17} & \cellcolor[HTML]{E6F4FF}\textbf{100.0\%} & \cellcolor[HTML]{E6F4FF}\textbf{100.0\% $\pm$ 0.0\%} & \cellcolor[HTML]{E6F4FF}\textbf{2.104 $\pm$ 2.224} & \cellcolor[HTML]{E6F4FF}\textbf{0.007 $\pm$ 0.004} \\
\addlinespace
\multirow{3}{*}{``Snake''}         & M1 & 95.03 $\pm$ 20.01         & 60.0\% & 87.5\% $\pm$ 17.7\% & 4.272 $\pm$ 2.599   & 0.038 $\pm$ 0.030 \\
                                        & M2 & \textbf{83.19 $\pm$ 20.89} & \textbf{88.9\%} & \textbf{97.2\% $\pm$ 8.3\%} & \textbf{1.459 $\pm$ 0.816} & 0.027 $\pm$ 0.013 \\
                                        & \cellcolor[HTML]{E6F4FF}M3 & \cellcolor[HTML]{E6F4FF}88.16 $\pm$ 31.59         & \cellcolor[HTML]{E6F4FF}80.0\% & \cellcolor[HTML]{E6F4FF}92.5\% $\pm$ 16.9\% & \cellcolor[HTML]{E6F4FF}1.836 $\pm$ 1.103   & \cellcolor[HTML]{E6F4FF}\textbf{0.023 $\pm$ 0.007} \\
\addlinespace
\multirow{3}{*}{``Frame''}         & M1 & 88.61 $\pm$ 15.61         & 88.9\% & 91.7\% $\pm$ 25.0\% & 5.394 $\pm$ 2.572   & 0.034 $\pm$ 0.019 \\
                                        & M2 & \textbf{79.98 $\pm$ 14.42} & 88.9\% & 94.4\% $\pm$ 16.7\% & 3.194 $\pm$ 2.333   & 0.023 $\pm$ 0.016 \\
                                        & \cellcolor[HTML]{E6F4FF}M3 & \cellcolor[HTML]{E6F4FF}88.39 $\pm$ 22.93         & \cellcolor[HTML]{E6F4FF}\textbf{100.0\%} & \cellcolor[HTML]{E6F4FF}\textbf{100.0\% $\pm$ 0.0\%} & \cellcolor[HTML]{E6F4FF}\textbf{2.073 $\pm$ 1.641} & \cellcolor[HTML]{E6F4FF}\textbf{0.019 $\pm$ 0.015} \\
\addlinespace
\multirow{3}{*}{Overall}                & M1 & 83.71 $\pm$ 19.73         & 55.6\% & 73.6\% $\pm$ 32.6\% & 4.947 $\pm$ 2.619   & 0.033 $\pm$ 0.022 \\
                                        & M2 & \textbf{78.45 $\pm$ 16.81} & 69.7\% & 83.1\% $\pm$ 28.5\% & 2.399 $\pm$ 1.855   & 0.021 $\pm$ 0.013 \\
                                        & \cellcolor[HTML]{E6F4FF}M3 & \cellcolor[HTML]{E6F4FF}79.18 $\pm$ 26.45         & \cellcolor[HTML]{E6F4FF}\textbf{75.0\%} & \cellcolor[HTML]{E6F4FF}\textbf{85.4\% $\pm$ 27.6\%} & \cellcolor[HTML]{E6F4FF}\textbf{1.927 $\pm$ 1.610} & \cellcolor[HTML]{E6F4FF}\textbf{0.016 $\pm$ 0.012} \\
\addlinespace
\bottomrule
\end{tabular}
\vspace{-0.10in}
\end{table*}

\subsubsection{Experimental Hypothesis}
\label{sec:Hypothesis}
We formulate the following primary hypotheses for our study: ($\pmb{H_1}$) the proposed system increases the quality of the assembly for participants; ($\pmb{H_2}$) the proposed system achieves higher success rates in robotic assembly tasks compared to standard teleoperation methods; ($\pmb{H_3}$) the proposed system is intuitive and imposes a lower cognitive workload on users during task execution; and ($\pmb{H_4}$) the proposed system show improved usability compare to standard teleoperation methods.

\subsubsection{Participant Study}
We evaluated \textsc{SUBTA} (M3) in comparison with the motion-support mode (M2) and standard teleoperation (M1) under the experimental setup described in \Cref{sec:exp_set_up}. All procedures complied with relevant ethical guidelines and regulations. Participants were eligible if they were (1) at least 18 years old, (2) free of known physical or cognitive impairments, and (3) had no history of musculoskeletal disorders. Examples from the study are shown in \Cref{fig:subject_demo}. Prior to testing, each participant completed a practice session with all modes. During the experiment, participants performed three randomly selected tasks out of four under each mode. The order of modes was randomized, and participants advanced once they reported being comfortable.

We recruited $N=12$ adult participants. 12 participants (11 male, 1 female) took part in the study, with ages ranging from 24 to 44 years (mean 33). On average, participants reported low gaming frequency (2.8/7) and moderate experience with robot teleoperation (3.7/7). Based on pilot/observed effect sizes, an a priori power analysis for our planned \emph{pairwise} contrasts against the baseline mode (M1) indicated that, with two-sided $\alpha=.05$, approximately 7--11 participants would achieve 80\% power for the subjective outcomes (e.g., TLX Mental, SUS total)~\cite{bausell2002power}. In addition, the objective \emph{accuracy} outcomes (position/orientation) error exhibited large effects when compared to M1, for which our within-subject design provides high sensitivity; thus our final sample of $N=12$ was adequate for these primary comparisons.

\section{Results and Discussion}
In what follows, we compare the proposed \textsc{SUBTA} system against two baseline modes using both objective and subjective outcomes, and evaluate how these results bear on the hypotheses in \Cref{sec:Hypothesis}. The definition and computation of each metric are provided in the following sections. Linear mixed-effects model results are summarized in \Cref{tab:hypothesis_pvalues}; note that the pose accuracy is evaluated using position errors and orientation errors, shown in the first and second rows. Overall, \textsc{SUBTA} (M3) outperforms standard teleoperation (M1): it achieves significantly higher assembly accuracy, lower reported mental workload (TLX--Mental), and higher SUS scores. Therefore, $\pmb{H_1}$, $\pmb{H_3}$, and $\pmb{H_4}$ are supported. For $\pmb{H_2}$, although the M3 vs. M1 comparison in success rate emphasizes practical rather than statistical significance, M3 achieves a substantial absolute gain in completion (55.6\% $\rightarrow$ 75\%), so we regard $\pmb{H_2}$ as supported as well.

\subsection{Quantitative Analysis}
\label{sec:objective}
The quantitative results are summarized in \Cref{tab:mode_summary_per_task_multirow}. Task progress is measured as the number of correctly placed blocks relative to the total blocks in the assembly. We treat the first placed anchor block as defining the global reference frame for that trial; all subsequent target poses are computed by applying the known relative transforms from $\mathcal{G}_T$ to that anchor. Position and orientation errors are computed by comparing user-placed block poses against the ground-truth poses.
Linear mixed-effects analysis revealed that \textsc{SUBTA} (M3) significantly outperformed standard teleoperation (M1) in position accuracy.
A linear mixed-effects model revealed significant improvements in position accuracy ($t(99)=7.88$, $p<0.001$, $d=1.18$) and orientation accuracy ($t(99)=12.67$, $p<0.001$, $d=1.75$), as well as a significant reduction in mental demand ($t(99)=4.65$, $p=0.002$, $d=1.34$.
M3 delivered up to a two-fold gain in pose accuracy relative to M1. Notably, M2 provides advantages in specific tasks, for example, in the “snake” assembly, participants achieved 97.2\% completion progress. M2 also yields slightly faster overall completion times (79.18 sec $\rightarrow$ 78.45 sec) than M3, suggesting that motion-level assistance can facilitate quicker grasping and placement compared to baseline teleoperation in general. 


\begin{figure}[tpb]
  \centering
  \includegraphics[width=\columnwidth]{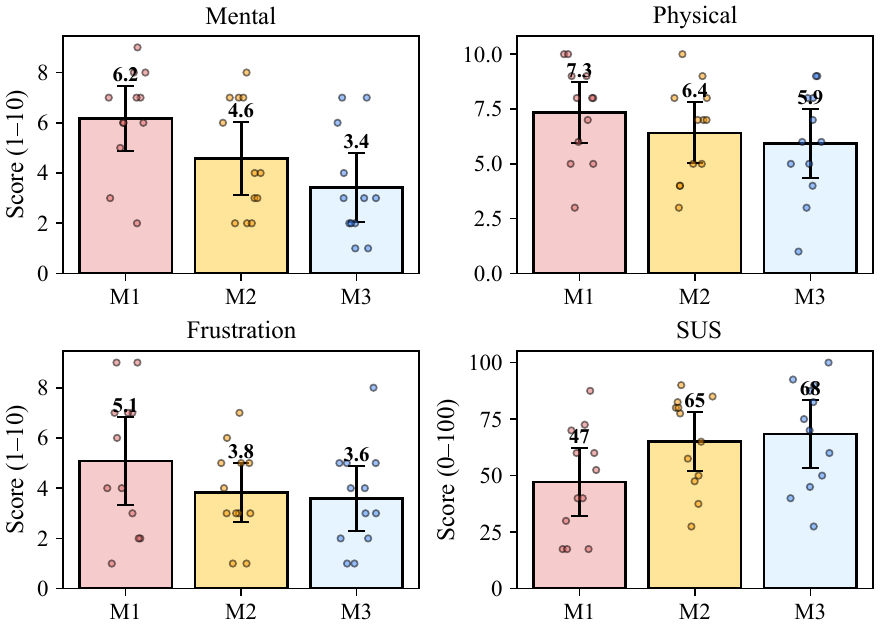}
  \caption{NASA-TLX ratings for mental demand, physical demand, and frustration, along with SUS scores across the three modes.}
  \label{fig:nasa}
  \vspace{-0.15in}
\end{figure}
\begin{table*}[ht]
\centering
\caption{Post-experiment questionnaire items for M2 and M3 on a 10-point Likert scale (1 = strongly disagree, 10 = strongly agree).}
\label{tab:survey_items}
\begin{tabular}{p{2cm}p{6cm}p{6cm}}
\toprule
\textbf{Category} & \textbf{M2 Items} & \textbf{M3 Items} \\
\midrule
Visual 
& (1) Placement surface visualization reduced ambiguity \newline
  (2) Placement surface visualization increased my confidence
& (1) Predicted target block visualization reduced ambiguity \newline
  (2) Predicted target block visualization increased my confidence \newline
  (3) Visualization was accurate and trustworthy \newline
  (4) Visualization was uncluttered \\
\hline
Motion 
& (3) Hand snapping made grasping easier \newline
  (4) Object snapping made placement easier \newline
  (5) Snapping increased my confidence \newline
  (6) Interventions were predictable and transparent
& Same as M2 (items 3--6) \\
\hline
Agency 
& (7) I always felt in control \newline
  (8) Easy to override/disengage snapping
& Same as M2 (items 7--8) \\
\bottomrule
\end{tabular}
\vspace{-0.15in}
\end{table*}

\subsection{Survey Responses}
\label{sec:subjective}
As shown in \Cref{fig:nasa}, we summarize subjective workload and usability across the three modes. TLX dimensions (Mental, Physical, Frustration) are scored on 1--10 (lower is better), and SUS on 0--100 (higher is better). We computed the sample mean $\bar{x}$ and standard deviation $s$ and formed a
$95\%$ confidence interval (CI) for the population mean using a $t$-distribution:
$\bar{x}\pm t_{0.975,\,n-1}\,\frac{s}{\sqrt{n}}$, where $n$ is the sample size. These intervals are shown as the vertical error bars in the figure.
The jittered points indicate the responses from individual participants.

\textbf{Mental demand.} We observe the means decrease monotonically ($\mathrm{M1} > \mathrm{M2} > \mathrm{M3}$), with visibly tighter CIs for M3. Paired-samples comparisons relative to M1 indicate substantial reductions in perceived cognitive effort: M3 vs.\ M1, $p{<}0.01$; M2 vs.\ M1, $p{<}0.01$; M3 vs.\ M2, not significant ($p{>}0.05$). These results suggest that SUBTA markedly lowers mental workload compared to standard teleoperation.

\textbf{Physical demand.} Significant differences were also observed for \emph{physical demand} across the three modes. The linear mixed-effects model result shows M3 vs.\ M1: $p{<}0.01$ and M2 vs.\ M1: $p\approx 0.05$. Participant-level points cluster at lower values under M3, indicating more consistently reduced effort. Overall, both M2 and M3 reduce physical demand, with M3 providing the greatest reduction.

\textbf{Frustration.} The same pattern holds for frustration: M3 has the lowest mean and a tighter distribution, indicating less irritation or stress than the standard teleoperation, while M2 remains intermediate between M1 and M3.

\textbf{Usability.} SUS (0--100) measures perceived usability. The mean scores rise monotonically ($\mathrm{M1}<\mathrm{M2}<\mathrm{M3}$) from all participants. The linear mixed-effects model result shows the same pattern: M3 vs.\ M1, $p\approx .001$; M2 vs.\ M1, $p\approx .001$; M3 vs.\ M2, not significant ($p{>}0.05$). Thus, both assisted modes are rated more usable than standard teleoperation mode, and M3 has the highest mean, though not reliably higher than M2. For context, a SUS of about 68 is commonly treated as ``average'' usability~\cite{lewis2018system}; \textsc{SUBTA} (M3) exceeds this benchmark and clearly improves over standard teleoperation (M1).

Overall, participants report \emph{lower} mental/physical workload and frustration and \emph{higher} usability in SUBTA (M3) compared to standard teleoperation (M1), with motion-support (M2) providing intermediate benefits.


\subsection{Visual, Motion, and Agency Feedback}
\begin{figure}[tbp]
  \centering
  \includegraphics[width=\columnwidth]{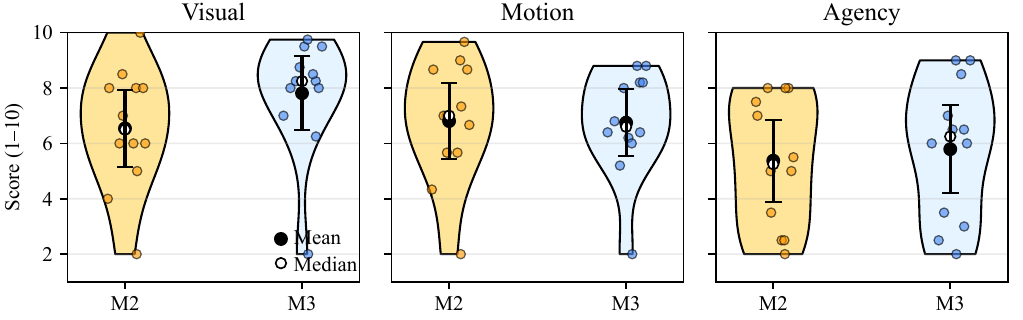}
  \caption{Full distributions with mean, median, and 95\% confidence intervals for the post-experiment comparison of M2 and M3.}
  \label{fig:post}
  \vspace{-0.15in}
\end{figure}

The post-experiment questions are summarized in \Cref{tab:survey_items}. We analyzed post-experiment ratings for the M2 and M3 across three domains, \emph{Visual}, \emph{Motion}, and \emph{Agency}. The distributions of the responses are visualized in \Cref{fig:post}.

\textbf{Visual feedback}: M3 provides substantially stronger visual support than M2. On the composite visual score, the mean increased from $\approx6.5$ in M2 to $\approx7.8$ in M3 ( shown in the figure), with a medium within-subject effect (paired trend: $p\approx.08$). Participants thus rated the \emph{predicted-target-block} visualization as clearer and more helpful than the \emph{placement-surface} visualization in M2. This aligns with our broader results: M3’s visual guidance reduces ambiguity about what to act on and boosts confidence in teleoperation. Practically, these findings argue for predictive visual cues that are accurate, uncluttered, and specific to the intended target.

\textbf{Motion feedback}: Composite ratings for snapping behaviors and intervention predictability were comparable in M2 and M3 (means $\approx6.8$ vs.\ $\approx6.8$, $p\approx.89$). This suggests that the \emph{motor-level assistance} (e.g., hand/object snapping, placement assistance, and transparency of interventions) was already effective in M2 and did not markedly change in perceived helpfulness when moving to M3. A design implication is that further gains in motion-support feedback may require more adaptive triggering when a snap occurs.

\textbf{Agency feedback}: Agency ratings (``felt in control'' and ``easy to override/disengage'') show a small improvement from M2 to M3 (means $\approx5.4\rightarrow5.8$, paired trend $p\approx.12$). Although not statistically significant with $N = 12$, the shift and the violin distributions indicate that users did not feel overruled by M3’s additional assistance, instead, the users reported M3 feels \emph{slightly greater} in control. Together with the predictability items, this supports the view that M3’s interventions were understandable and that override remained accessible.

Taken together, these subjective data indicate that \textsc{M3}’s main added value comes from its \emph{visual} guidance: clearer target specification and improved user confidence. Perceived \emph{motion} assistance is already strong in M2 and remains comparable in M3, while \emph{agency} shows a modest upward shift in favor of M3. In combination with previous objective results, the evidence supports \textbf{SUBTA (M3) as the preferable mode: it leverages predictive visual feedback without eroding user control or transparency}.

\section{Conclusion}
We presented \textsc{SUBTA}, a supported teleoperation framework that integrates learned task and intention estimation, scene-graph--based task planning, and context-dependent motion support for bimanual assembly. In the user study, \textsc{SUBTA} achieved significantly higher pose accuracy (position and orientation), lower NASA--TLX mental load, and higher usability scores (SUS) compared to standard teleoperation. It also improved success rates in absolute terms (55.6\% $\rightarrow$ 75\%). The study further demonstrated the benefits of systematically combining visual and motion support, showing that clear target specification improved user confidence. Future work will include larger-scale user studies and the development of adaptive policies to enhance perceived motion support and user agency while maintaining system predictability.






\bibliographystyle{unsrt}
\bibliography{reference}

\end{document}